\pdfoutput=1
\documentclass[10pt,twocolumn,letterpaper]{article}
\usepackage{iccv}
\usepackage{times}
\usepackage{epsfig}
\usepackage{graphicx}
\usepackage{amsmath}
\usepackage{amssymb}
\newcommand{\Jia}[1]{{\color{black}#1}\normalfont}
\usepackage[accsupp]{axessibility}

\usepackage[pagebackref=true,breaklinks=true,letterpaper=true,colorlinks,bookmarks=false]{hyperref}

\iccvfinalcopy 

\def\iccvPaperID{1241} %

\begin{document}
\def\etc{\emph{etc}\onedot} \def\vs{\emph{vs}\onedot}

\title{Learning Facial Representations from the Cycle-consistency of Face}

\author{Jia-Ren Chang \qquad Yong-Sheng Chen \qquad Wei-Chen Chiu \\
National Yang Ming Chiao Tung University, Hsinchu, Taiwan \\
{\tt\small $\{$followwar.cs00g, yschen, walon$\}$@nctu.edu.tw}\\
}

\maketitle

\begin{abstract}
Faces manifest large variations in many aspects, such as identity, expression, pose, and face styling.
Therefore, it is a great challenge to disentangle and extract these characteristics from facial images, especially in an unsupervised manner. 
In this work, we introduce cycle-consistency in facial characteristics as free supervisory signal to learn facial representations from unlabeled facial images.
The learning is realized by superimposing the facial motion cycle-consistency and identity cycle-consistency constraints.
The main idea of the facial motion cycle-consistency is that, given a face with expression, we can perform de-expression to a neutral face via the removal of facial motion and further perform re-expression to reconstruct back to the original face.
The main idea of the identity cycle-consistency is to exploit both de-identity into mean face by depriving the given neutral face of its identity via feature re-normalization and re-identity into neutral face by adding the personal attributes to the mean face.
At training time, our model learns to disentangle two distinct facial representations to be useful for performing cycle-consistent face reconstruction. 
At test time, we use the linear protocol scheme for evaluating facial representations on various tasks, including facial expression recognition and head pose regression.
We also can directly apply the learnt facial representations to person recognition, frontalization and image-to-image translation.
Our experiments show that the results of our approach is competitive with those of existing methods, demonstrating the rich and unique information embedded in the disentangled representations.
Code is available at \url{https://github.com/JiaRenChang/FaceCycle}.
\end{abstract}

\section{Introduction}
\begin{figure}
    \centering
    \includegraphics[clip, trim= 0.0cm 15.6cm 14.6cm 0.0cm ,width=\columnwidth]{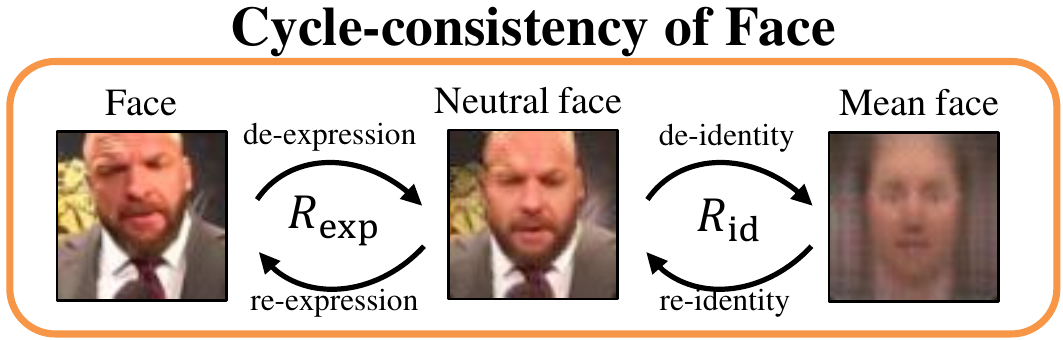}
    \caption{We propose an unsupervised framework based on cycle-consistency for learning face disentanglement. 
    We define that all the variations between a face and its corresponding neutral face of the same identity as \textit{expression}.
    Similarly, all the variations between a neutral face and the global mean face are defined as \textit{identity}.   
    The input face is sequentially deprived of expression ($R_{\mathrm{exp}}$) and identity ($R_{\mathrm{id}}$) representations by networks to become the neutral face and mean face, respectively, which can be transformed back to the original face in reverse order.
}
    \label{fig:multitask}
\end{figure}
 
Face perception is vital for human beings and is also essential in the field of computer vision. 
Neuroimaging studies of both human and monkey~\cite{hasselmo1989role,haxby2000distributed,winston2004fmri} reveal the neuroanatomical dissociation between expression and identity representations in face perception. 
Their findings suggest that these facial characteristics are processed in different brain areas. 
With the renaissance of deep learning in recent years, computer vision research field follows this thread of thinking and progresses in the direction of disentangling the face characteristics into separated low-dimensional latent representations, such as identity~\cite{tran2017disentangled}, expression~\cite{yang2018facial,zhang2018joint}, shape/appearance~\cite{shu2018deforming,xing2019unsupervised}, intrinsic images~\cite{shu2017neural}, and fine-grained attributes (age, gender, wearing glasses, etc.)~\cite{shen2017learning}.

Several supervised methods have been proposed to disentangle face characteristics for image manipulation by conditioning generative models on a pre-specified face representations, including landmarks~\cite{zakharov2019few}, action units~\cite{pumarola2018ganimation}, or facial attributes~\cite{lu2018attribute}. 
Particularly, these methods are able to manipulate faces while preserving the identity. 
Other studies incorporate head pose information to disentangle pose-invariant representations for robust identity~\cite{tran2017disentangled}/expression~\cite{zhang2018joint} recognition. 
Provided with neutral face, moreover, de-expression residue learning~\cite{yang2018facial} can facilitate the model to learn identity-invariant expression representations for performing facial expression recognition.

The 3D Morphable Model (3DMM)~\cite{blanz1999morphable,chu20143d} for face shape modeling incorporates a similar thinking of dissociation for expression and identity.
The most widely-used form of 3DMM is that a face shape ${\bf S}$ is a linear combination of mean shape $\Bar{\bf S}$ and identity and expression vectors (${\bf z}_{\textrm{id}}, {\bf z}_{\textrm{exp}}$): ${\bf S} = \Bar{\bf S} + {\bf A}_{\textrm {id}}{\bf z}_{\textrm{id}} + {\bf A}_{\textrm{exp}}{\bf z}_{\textrm{exp}}$, where ${\bf A}_{\textrm{id}}$ and ${\bf A}_{\textrm{exp}}$ are the identity and expression PCA bases, respectively. 
Jiang \etal~\cite{Jiang_2019_CVPR} introduce a variational autoencoder approach for learning latent representations of expression mesh and identity mesh in the framework of 3DMM. 
However, they provided strong supervision for the disentanglement of identity and expression representations, including ground truths of shape meshes for expression, identity, and the mean face~\cite{Jiang_2019_CVPR}. 
It is difficult to generalize such methods to 2D facial images without being given any ground truth.

In addition to the aforementioned works which are mostly based on supervised learning, recently a few studies begin to exploit the unsupervised learning framework to disentangle facial characteristics~\cite{lu2020self,wiles2018self,wiles2018x2face,xing2019unsupervised}. 
These methods focus on extracting a part of facial characteristics. 
\Jia{For example, FAb-Net~\cite{wiles2018self} learns representations that encode information about pose and expression, \cite{lu2020self, wiles2018self} introduce frameworks to learn representations for action unit detection, and }
Zhang \etal~\cite{zhang2018unsupervised} propose an autoencoder to locate facial landmarks. 
Some unsupervised methods~\cite{shu2018deforming,xing2019unsupervised} attempt to separate two independent representations of face images, including shape and appearance. 
However, these unsupervised methods can only disentangle a part of information of facial images, but not yet investigate a more general generative procedure of a human face, that is, simultaneous disentanglement of expression and identity representations for a wider usage.

In this paper, we propose a novel framework that is able to simultaneously disentangle expression and identity representations from 2D facial images in an unsupervised manner. Particularly, the definition of the \textbf{expression} factor in our proposed method contains all the variations between an arbitrary face image and its corresponding neutral face of the same identity, including the facial expression and head pose. While for the \textbf{identity} factor, we define it to contain all the variations between a neutral face and the global mean face, including the facial identity and other subject-specific features such as hair style, age, gender, beard, glasses, \textit{etc}. Based on these definitions, we propose two novel cycle-consistency constraints to drive our model learning, as illustrated in Figure~\ref{fig:multitask}.

The first cycle-consistency constraint stems from the idea of action unit~\cite{ekman1978facial} in which the head poses and facial expressions are the results of the combined and coordinated action of facial muscles.
Therefore, the head poses and expression can be treated as the optical flow~\cite{mase1991recognition} between a neutral face and any face of the same identity. 
To this end, a decoder is trained to learn the optical flow field of the input face \textit{without} the ground truth neutral face. 
This is achieved by applying the proposed idea called \textit{\textbf{facial motion cycle-consistency}}, which is able to perform both the \textbf{de-expression} and \textbf{re-expression} operations.

The second cycle-consistency constraint originates from Eigenfaces~\cite{sirovich1987low}, in which a facial image is represented by adding a linear combination of eigenfaces to the mean face, suggesting that the face identity is embedded in the linear combination of eigenfaces. 
Instead of representing the identity as the residue of neutral facial image relative to the mean face~\cite{sirovich1987low}, we model the adding and depriving of identity as a renormalization procedure, analogues to the feed-forward style transfer tasks~\cite{huang2017arbitrary}.
To this end, decoders are trained to learn the renormalized features \textit{without} the ground truth mean face.
This is achieved by applying the proposed idea called \textit{\textbf{identity cycle-consistency}}, which is able to perform identity deprivation as \textbf{de-identity} and the identity styling as \textbf{re-identity}.

The main contributions
of our work are summarized as follows: 
\begin{itemize}
\item We propose a novel framework for unsupervised learning of facial representations from a single facial image, based on the novel ideas of facial motion cycle-consistency and identity cycle-consistency.

\item The disentangled expression and identity features obtained by our proposed method can be easily utilized for various downstream tasks, 
such as facial expression recognition, head pose regression, person recognition, frontalization, and image-to-image translation.

\item We demonstrate that the performance of the learned representations in different downstream tasks is competitive with the state-of-the-art methods.
\end{itemize}

\section{Unsupervised Learning of Facial Representations}
\begin{figure*}[htbp]
    \centering
    \includegraphics[clip, trim= 1.4cm 8.8cm 4.1cm 4.6cm ,width=\textwidth]{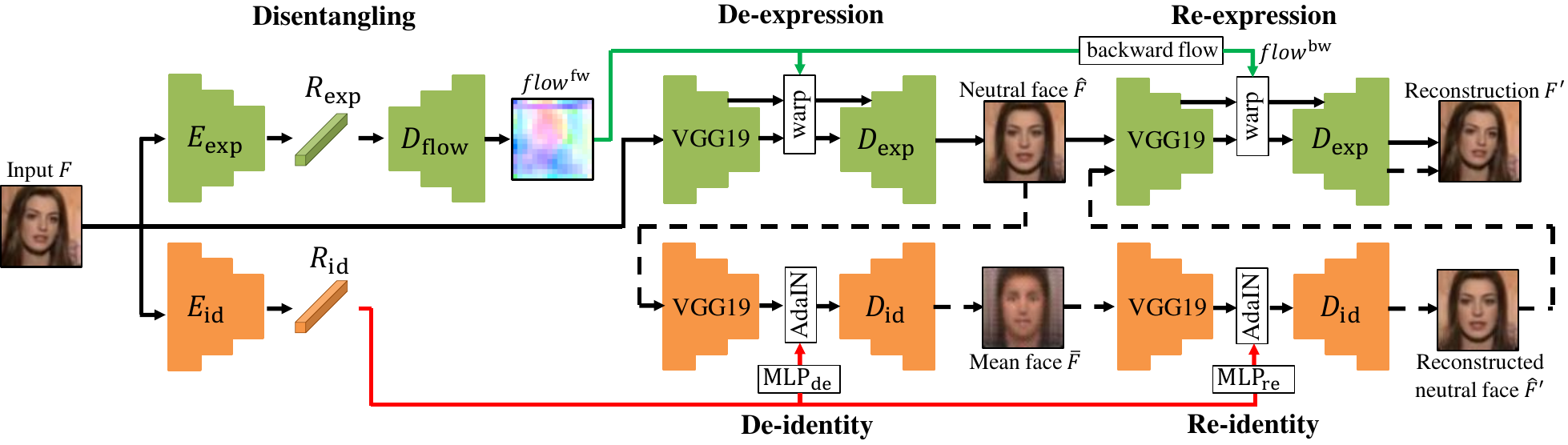}
    \caption{Overall architecture of the proposed model. The $E_{\textrm{exp}}$ and $E_{\textrm{id}}$ are trained to extract expression and identity representations respectively by using our unsupervised disentangling method. 
    By exploring the disentangled representations, networks $D_{\textrm{flow}}$, $D_{\textrm{exp}}$, MLPs and $D_{\textrm{id}}$ are trained to generate the representation-removed images, the neutral face $\hat{F}$ and the mean face $\bar{F}$, and to reconstruct the representation-added images, the input face $F^{'}$ and the neutral face $\hat{F}^{'}$.
    \Jia{Please note that the proposed method needs two images to train the model as described in Sec. \ref{sec:fmcc} and \ref{sec:idcc}, and we only show a single image forwarding here for simplicity.}
    }
    \label{fig:arch}
\end{figure*}

As motivated previously, in this paper we aim at disentangling the identity and expression representations from a single facial image. 
Our proposed method is mainly based on an important assumption that: a facial image $F$, from high-level perspective, can be decomposed as follows:
 \begin{equation}
 \begin{split}
      F = \bar{F} + \textrm{id} + \textrm{exp} = \hat{F} + \textrm{exp} \, ,   
 \end{split}
 \end{equation}
where $\bar{F}$ is the global mean face shared among all the faces, \textbf{id} and \textbf{exp} are the identity and expression factors respectively, and $\hat{F}$ is the neutral face of a particular identity specified by \textbf{id}. Therefore, our proposed model is trained to learn the expression and identity representations, denoted as $R_{\textrm{exp}}$ and $R_{\textrm{id}}$ respectively, for indicating the facial characteristics of facial images. 
%
We introduce four processes based on cycle-consistency for learning these representations, as shown in Figure~\ref{fig:multitask}:
\begin{itemize}
\item \textbf{de-expression.} We define the \textit{de-expression} as removing $R_{\textrm{exp}}$ from the input facial image $F$, in which we can obtain the neutral face $\hat{F}$ accordingly.
\item \textbf{re-expression.} 
The \textit{re-expression} is defined as assigning $R_{\textrm{exp}}$ to the neutral face $\hat{F}$ for reconstructing face with expressions $F^{'}$.
\item \textbf{de-identity.} 
We define \textit{de-identity} as an operation for removing $R_{\textrm{id}}$ from the input neutral face $\hat{F}$ in order to obtain the mean face $\bar{F}$.
\item \textbf{re-identity.} 
The \textit{re-identity} is defined as the process of recovering the neutral face $\hat{F}^{'}$ back from the mean face $\bar{F}$ according to $R_{\textrm{id}}$. 
\end{itemize}
As illustrated in Figure~\ref{fig:arch}, the overall architecture of our proposed model consists of
two encoders ($E_{\textrm{exp}}$ and $E_{\textrm{id}}$) for extracting expression and identity representations respectively, and two decoders ($D_{\textrm{exp}}$, $D_{\textrm{id}}$) for learning nonlinear mapping functions of the aforementioned four processes.
In the following, we detail the proposed unsupervised learning method to disentangle expression and identity representations.

\begin{figure*}
    \centering
    \includegraphics[clip, trim= 1.6cm 5.9cm 2.5cm 7.6cm,width=0.95\textwidth]{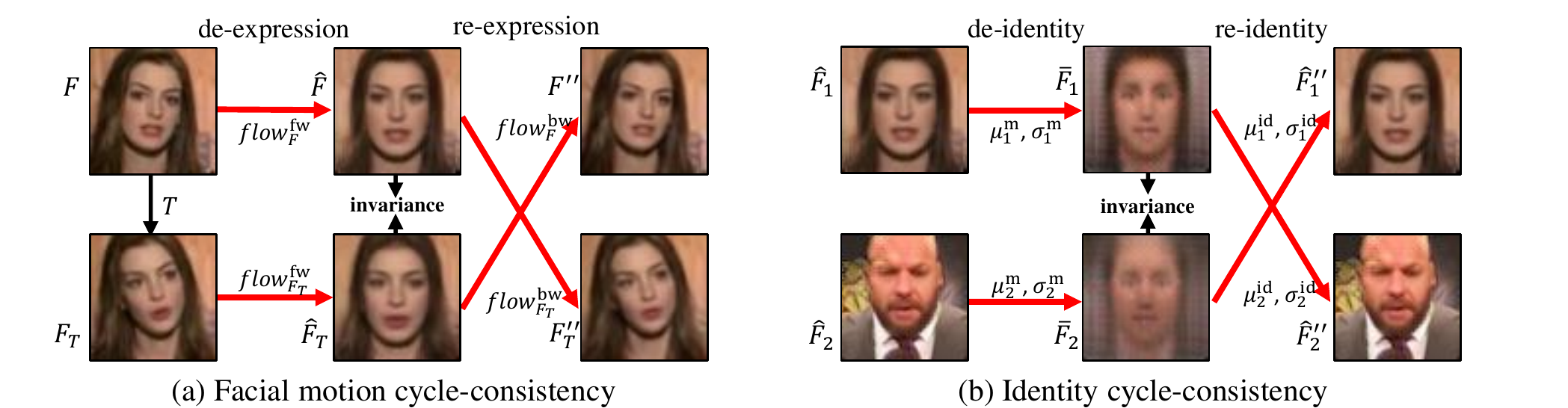}
    \caption{Illustration of (a) facial motion cycle-consistency for learning expression representations, and (b) identity cycle-consistency for learning identity representations.}
    \label{fig:deexp}
\end{figure*}


\subsection{Expression Representation}
\label{exprep}
We start with the introduction of facial motion cycle-consistency in the following.
We denote that the expression representations $R_{\textrm{exp}}$ are learned by an encoder $E_{\textrm{exp}}$ from the input face image $F$:
\begin{equation}
    R_{\textrm{exp}}=E_{\textrm{exp}}(F) \, .    
\end{equation}
As the idea described in the previous section, we model a facial expression as the optical flow field between the neutral face and the face with expression.
Therefore, the forward ($F\to \hat{F}$) optical flow field $flow^{\textrm{fw}} \in \mathbb{R}^{2\times H \times W}$, where $H$ and $W$ is the height and width of the input image, is learned by the decoder $D_{\textrm{flow}}$ from expression representations.
Moreover, according to the well-known \textit{forward-backward flow consistency}~\cite{black1996robust,hur2017mirrorflow}, we can compute the backward ($\hat{F}\to F$) optical flow field $flow^{\textrm{bw}}$ according to $flow^{\textrm{fw}}$, which basically is inverse $flow^{\textrm{fw}}$ by a warp function $\mathcal{W}$:
\begin{equation}
\label{flowcons}
\begin{split}
    flow^{\textrm{fw}}  =& D_{\textrm{flow}}(R_{\textrm{exp}}) \, ,\\ 
    flow^{\textrm{bw}}  =& - \mathcal{W}(flow^{\textrm{fw}}, flow^{\textrm{fw}}) \, .   
\end{split}
\end{equation}
We use bilinear interpolation to implement the warping operation $\mathcal{W}$ as in \cite{sun2018pwc}.
By using the forward optical flow field $flow^{\textrm{fw}}$ we can warp $F$ pixel-wisely to obtain an intermediate facial image, denoted as $\tilde{F}$. Followed by using the corresponding backward optical flow field $flow^{\textrm{bw}}$, we are able to warp back from $\tilde{F}$ to reconstruct $F$.
This procedure straightforwardly lead to a reconstruction loss $\mathcal{L}_{\textrm{flow}}$ which is defined as:
\begin{equation}
    \mathcal{L}_{\textrm{flow}} = | F- \mathcal{W}(\mathcal{W}(F, flow^{\textrm{fw}}),flow^{\textrm{bw}})| \, .
\end{equation}

Furthermore, we exploit a general image feature extraction to represent a face image, that is, the coarse-to-fine feature maps $feat_F$ obtained from layers \texttt{conv2$\_$1}, and \texttt{conv3$\_$1} of VGG19 network pre-trained on ImageNet~\cite{simonyan2014very}.
Given a forward flow field $flow^{\textrm{fw}}$, we simply use the bilinear interpolation function $d_s(\cdot)$ to obtain $d_s(flow^{\textrm{fw}})$ of the size equal to $feat_F$.
The \textbf{de-expression} is then achieved by first warping $feat_F$ with $d_s(flow^{\textrm{fw}})$ and then adopting a decoder $D_{\textrm{exp}}$ to generate neutral face image $\hat{F}$:
\begin{equation}
    \hat{F}=D_{\textrm{exp}}(\mathcal{W}(feat_F, d_s(flow^{\textrm{fw}}))) \, .
\end{equation}
Moreover, we argue that the image features $feat_{\hat{F}}$ of a neutral face obtained by VGG19 could be warped back via the downsampled backward flow $d_s(flow^{\textrm{bw}})$ and then be fed into the decoder $D_{\textrm{exp}}$ for reconstructing a face with expression, denoted as $F^{'}$, which ideally should be identical to the original face $F$.
This process is exactly the \textbf{re-expression}:
\begin{equation}
\label{exprecon}
  F^{'}=D_{\textrm{exp}}(\mathcal{W}(feat_{\hat{F}}, d_s(flow^{\textrm{bw}}))) \, .  
\end{equation}

\subsection{Facial Motion Cycle-Consistency: Invariance for Learning Expression Representation}\label{sec:fmcc}
The change on a face image $F$ caused by a facial motion can be expressed in terms of a spatial image transformation $T$, where we denote the corresponding face image with different motion but the same identity as $F_T$.
As both $F$ and $F_T$ are with the same identity, their corresponding neutral faces should be also identical.
That is, their decoded neutral faces after performing de-expression are \textbf{invariant} to each other, which leads to the constraint: \begin{equation}
\label{expinv}
    \hat{F} = \hat{F_T} \, .
\end{equation}
Following the concept of this invariance, we should be able to apply the re-expression operation on $feat_{\hat{F_T}}$ (the features for the decoded neutral face of $F_T$) via the downsampled backward flow $d_s(flow^{\textrm{bw}}_{F})$ of $F$ (related to the expression of $F$) to reconstruct a face image denoted as $F^{''}=D_{\textrm{exp}}(\mathcal{W}(feat_{\hat{F_T}}, d_s(flow^{\textrm{bw}}_{F})))$, which ideally is quite similar to the original $F$ due to the hypothesis that $feat_{\hat{F}}=feat_{\hat{F_T}}$ as $\hat{F}=\hat{F_T}$.
The similar story holds for performing re-expression on $feat_{\hat{F}}$ by $d_s(flow^{\textrm{bw}}_{F_T})$ to reconstruct $F^{''}_T=D_{\textrm{exp}}(\mathcal{W}(feat_{\hat{F}}, d_s(flow^{\textrm{bw}}_{F_T})))$, which is almost identical to $F_T$. The illustration of this invariance, also named as \textbf{facial motion cycle-consistency}, is shown in Figure~\ref{fig:deexp}(a). 

The reconstruction derived from the invariance 
(that is $F^{''}$ versus $F$ and $F_T^{''}$ versus $F_T$) 
builds up the objectives $\mathcal{L}_{\textrm{exp}}$ for learning the expression representations $R_{\textrm{exp}}$, where we utilize both the L1 loss and the perceptual loss~\cite{gatys2016image,johnson2016perceptual} to evaluate the error of reconstruction:
\begin{equation}
\begin{aligned}
    \mathcal{L}_{\textrm{exp}}(F, F_T) =&|F-F^{''}|+|F_T-F_T^{''}| \\
    +&\lambda(\Phi(F, F^{''})+\Phi(F_T, F^{''}_T)) \, ,
\end{aligned}
\end{equation}
where $\lambda$ is set to $0.05$ to balance the L1 and perceptual losses.
The perceptual loss is defined as $\Phi(F, F^{''})=\sum_{l}\lVert\mathit{\phi}^{l} (F)-\mathit{\phi}^{l}(F^{''})\rVert_2 + \sum_{l}\lVert\mathcal{G}(\mathit{\phi}^{l} (F))-\mathcal{G}(\mathit{\phi}^{l}(F^{''}))\rVert_2$. The function $\phi^{l}(\cdot)$ extracts VGG19 features from layer $l$, in which the \texttt{conv2\_1}, \texttt{conv3\_1}, and \texttt{conv4\_1} layers are used here. The function $\mathcal{G}(\cdot)$ calculates the Gram matrix of the feature map.

\subsection{Identity Representation}
In terms of identity representation $R_{\textrm{id}}$, we utilize the encoder $E_{\textrm{id}}$ to extract $R_{\textrm{id}}$ from the input face image $F$: 

\begin{equation}
    R_{\textrm{id}} = E_{\textrm{id}}(F) \, .
\end{equation}
Based on the idea described previously, we argue that the identity representation could be deprived from the neutral face to obtain the mean face. To implement the \textbf{de-identity} operation, we design a decoder $D_{\textrm{id}}$ to generate the mean face $\bar{F}$ from the modulated VGG features $feat_{\hat{F}}$ of a neutral face $\hat{F}$, which is similar to the idea of feature modulation idea proposed in the AdaIN paper~\cite{huang2017arbitrary}:
\begin{equation}
    \bar{F}=D_{\textrm{id}}(\frac{feat_{\hat{F}}-\mu({feat_{\hat{F}}})}{\sigma(feat_{\hat{F}})}\sigma^{\textrm{m}}+\mu^{\textrm{m}}) \, ,
\end{equation}
where $\mu(\cdot)$ and $\sigma(\cdot)$ are used to compute the mean and standard deviation respectively, and $\mu^{\textrm{m}}$ and $\sigma^{\textrm{m}}$ are learned from $R_{\textrm{id}}$ by the multi-layer perceptron $\textrm{MLP}_{\textrm{de}}$:
\begin{equation}
    \mu^{\textrm{m}}, \sigma^{\textrm{m}} = \textrm{MLP}_{\textrm{de}}(R_{\textrm{id}}) \, .
\end{equation}
Furthermore, the \textbf{re-identity} can be achieved in a similar manner but reversely with the decoder $D_{\textrm{id}}$:
\begin{equation}
    \hat{F^{'}}=D_{\textrm{id}}(\frac{feat_{\bar{F}}-\mu({feat_{\bar{F}}})}{\sigma(feat_{\bar{F}})}\sigma^{\textrm{id}}+\mu^{\textrm{id}})\, ,
\end{equation}
where the $\mu^{\textrm{id}}$ and $\sigma^{\textrm{id}}$ are also learned from $R_{\textrm{id}}$ but by another multi-layer perceptron $\textrm{MLP}_{\textrm{re}}$.


\subsection{Identity Cycle-Consistency: Invariance for Learning Identity Representation}\label{sec:idcc}
We hypothesize that the mean face is global for all the faces.
In other words, no matter starting from which neural face of any identity, we should always obtain the same mean face after performing de-identity operation.
Given the neutral faces $\hat{F_1}$ and $\hat{F_2}$ of different identities, we can derive the \textbf{invariance} related to identity as:
\begin{equation}
    \bar{F_1} = \bar{F_2} \, .
\end{equation}
Therefore, we should be able to reconstruct $\hat{F_1}$ by using its corresponding $\{\mu_1^{\textrm{id}}, \sigma_1^{\textrm{id}}\}$ to apply the re-identity operation on the mean face obtained from $\hat{F_2}$.
The result of this reconstruction is denoted as $\hat{F}_1^{''}$:
\begin{equation}
\label{idcons}
\hat{F}_1^{''}=D_{\textrm{id}}(\frac{feat_{\bar{F}_2}-\mu(feat_{\bar{F}_2})}{\sigma(feat_{\bar{F}_2})}\sigma^{\textrm{id}}_1+\mu^{\textrm{id}}_1) \, .
\end{equation}
Again, similar story holds to perform re-identity operation (with $\{\mu_2^{\textrm{id}}, \sigma_2^{\textrm{id}}\}$) on the mean face obtained from $\hat{F_1}$ to reconstruct $\hat{F_2}$.
We denote the reconstruction result as $\hat{F}_2^{''}$:
\begin{equation}
\label{idcons2}
\hat{F}_2^{''}=D_{\textrm{id}}(\frac{feat_{\bar{F}_1}-\mu(feat_{\bar{F}_1})}{\sigma(feat_{\bar{F}_1})}\sigma^{\textrm{id}}_2+\mu^{\textrm{id}}_2) \, .
\end{equation}
The illustration of this invariance related to identity representations, also named as \textbf{identity cycle-consistency}, is shown in  Figure~\ref{fig:deexp}(b).

As the way that $\mathcal{L}_{\textrm{exp}}$ is defined, the reconstruction derived from the invariance (that is, $\hat{F}_1^{''}$ versus $\hat{F}_1$ and $\hat{F}_2^{''}$ versus $\hat{F}_2$) leads to the objectives $\mathcal{L}_{\textrm{id}}$ for learning the identity representations $R_{\textrm{id}}$:
\begin{equation}
\begin{aligned}
    \mathcal{L}_{\textrm{id}}(\hat{F}_1, \hat{F}_2) =&|\hat{F}_1-\hat{F}_1^{''}|+|\hat{F}_2-\hat{F}_2^{''}| \\
    +& \lambda(\Phi(\hat{F}_1, \hat{F}_1^{''})+\Phi(\hat{F}_2, \hat{F}_2^{''})) \, .
\end{aligned}
\end{equation}
Moreover, we additionally introduce a margin loss $\mathcal{L}_{\textrm{m}}$ to constrain the mean face:
\begin{equation}
   \mathcal{L}_{\textrm{m}}(\bar{F}, \hat{F}) = \max({\left \| \bar{F}-\hat{F} \right \|
   }-\alpha, 0) \, , 
\end{equation}
where we set $\alpha=0.1$ in all experiments.
The main motivation behind this margin loss is that we would like to constrain the difference between the mean face and the neutral face to be within a margin. Otherwise the obtained mean face could potentially become an arbitrary image far from a face image.

\section{Experiments}
We report experimental results for a model trained on
the combination of VoxCeleb1~\cite{Nagrani17} and VoxCeleb2~\cite{Chung18b} from scratch.
The trained representations are evaluated on several tasks, including facial expression recognition, head pose regression, person recognition, frontalization, and image-to-image translation. 
Through various experiments, we show that the acquired representation generalizes to a range of facial image processing tasks.

\subsection{Training Procedures}
The facial motion cycle-consistency described in Section~\ref{sec:fmcc} involves an image pair of faces with different expressions/poses but of the same identity. Fortunately, this type of data can be easily available from the video recording of human faces, for instance, the video of interview or talk-show, which exists widely on the Internet nowadays. 
Given any two frames in this type of video clip of a person, we can easily obtain a pair of facial images showing different expressions. 
Therefore, we can take the advantage of this type of video sequences (as the dataset described in the following) and collect training data for learning both the expression and identity representations in an unsupervised manner.

\paragraph{Dataset.}

The proposed model is trained on the combination of VoxCeleb1~\cite{Nagrani17} and VoxCeleb2~\cite{Chung18b} datasets, in which both datasets are built upon videos of interviews.
VoxCeleb1 has in total 153,516 video clips of 1,251 speakers, while VoxCeleb2 has 145,569 video clips of 5,994 speakers.
Video frames were extracted at 6 fps, cropped to have faces shown in the center of frames, and then resized to the resolution of 64$\times$64.
We adopted VoxCeleb2 test dataset for visualizing the intermediate results of our disentanglement process.

\paragraph{Stage-wise Training Procedure.} We introduce a stage-wise training procedure for our model learning. There are two main stages for sequentially training different parts of the proposed model, in order to disentangle the expression and identity representations.\\
\noindent\textbf{-- Stage 1: training of $E_{\textrm{exp}}$, $D_{\textrm{flow}}$ and $D_{\textrm{exp}}$}\\
For training the subnetworks related to the de-expression and re-expression parts, as the green-shaded components shown in Figure~\ref{fig:arch}, the objectives of $\mathcal{L}_{\textrm{flow}}$ and $\mathcal{L}_{\textrm{exp}}$ are utilized to update $\{E_{\textrm{exp}},D_{\textrm{flow}},D_{\textrm{exp}}\}$. 
The transformation $T$ needed for the use of $\mathcal{L}_{\textrm{exp}}$ can be simply obtained by having the horizontal flipping (that is, $F_T$ is the horizontally flipped version of $F$) or taking any arbitrary pair of faces from different frames (that is, two faces of the same person shown at different times in a video).
We provide an ablation study in the supplementary materials.
\\
\noindent\textbf{-- Stage 2: training of $E_{\textrm{id}}$, $D_{\textrm{id}}$, $\textrm{MLP}_{\textrm{re}}$, and $\textrm{MLP}_{\textrm{de}}$}\\
For training the subnetworks related to the de-identity and re-identity parts, as the orange-shaded components shown in  Figure~\ref{fig:arch}, both the objectives of $\mathcal{L}_{\textrm{id}}$ and $\mathcal{L}_{\textrm{m}}$ are applied to update all of these subnetworks. 

\paragraph{Implementation Details.}
Our proposed model is implemented based on PyTorch framework and trained with the Adam optimizer ($\beta_1=0.5$, and $\beta_2=0.999$).
The batch size is set to 32 for all the training stages.
The initial learning rate is 0.00005 in the Stage 1 and 0.0001 in the Stage 2. The Stage 1 and Stage 2 are trained for 40 and 20 epochs respectively.
The learning rate is decreased by a factor of 10 at half of total epochs.
Moreover, both representation encoders (\textit{i.e.} $E_{\textrm{exp}}$ and $E_{\textrm{id}}$) adopt the same network architecture which is a 16-layer CNN.
We leverage a VGG-19 \cite{simonyan2014very} for the general feature extraction (denoted as VGG19 component in the Figure~\ref{fig:arch}), where the VGG-19 encoded facial features can be further passed through our decoders (\textit{i.e.} $D_{\textrm{exp}}$ or $D_{\textrm{id}}$) to generate new facial images.
The model architectures are detailed in the supplementary materials.

\paragraph{Baselines.}
We adopt the following baselines for making evaluations and comparisons in terms of the quality and representativeness of the extracted facial features:
\\
\noindent\textbf{-- HoG descriptor~\cite{dalal2005histograms}:}
We follow the same setting as in~\cite{li2017reliable}, where the facial images are first rescaled to the size of 100$\times$100, then the HoG feature of 3,240 dimensions is extracted for each image. 
\\
\noindent\textbf{-- LBP descriptor~\cite{ojala2002multiresolution}:}
Similar to the HoG descriptor, we follow the same setting as in~\cite{li2017reliable} to extract 1,450 dimensional LBP feature vector from each of the facial images which are resized to 100$\times$100.
\\
\noindent\textbf{-- MoCo~\cite{he2020momentum}:}
We adopt the state-of-the-art self-supervised representation learning method, MoCo, as a strong baseline for us to compare with. We follow the MoCo algorithm to train the feature extractor (in which its network architecture is the same as our encoders) based on the same training dataset as ours (\textit{i.e.} VoxCeleb1 and VoxCeleb2). The training runs for 40 epochs with SGD optimizer, batch size of 128, momentum 0.999, and 65,536 negative keys.
\\
\noindent\textbf{-- Self-supervisely learnt facial representations:}
\Jia{Three state-of-the-art self-supervised frameworks of facial representation learning \cite{li2019self, lu2020self, wiles2018self} are utilized to compare with our work.
We directly adopt the models officially released by their authors (which are all pretrained on the Voxceleb dataset) for experimenting the downstream tasks of expression classification and head pose regression.
Please note that we apply a linear protocol on their learnt features to have a fair comparison.}



\subsection{Intermediate Results of Our Model}
\label{NetOut}
\begin{figure}
    \centering
    \includegraphics[clip,trim= 0cm 9cm 13.9cm 0 cm,width=\columnwidth]{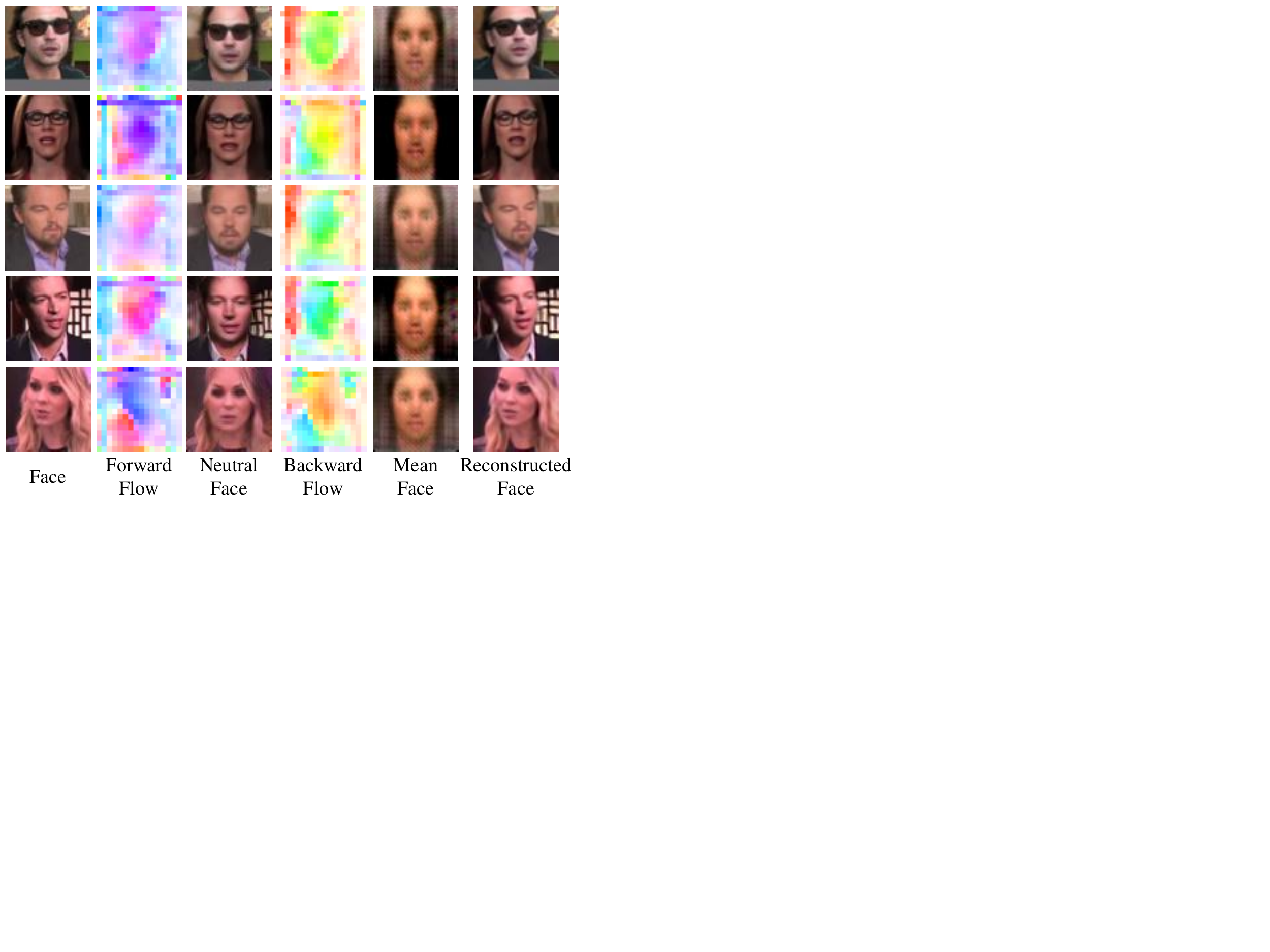}
    \caption{Visualization of intermediate results of our model. Input images are from the test set of the VoxCeleb2 dataset~\cite{Chung18b}.
    From left to right, the columns sequentially show the input faces, forward flow fields, neutral faces, backward flow fields, mean faces, and the faces reconstructed from the corresponding neutral ones.}
    \label{fig:output}
\end{figure}

Figure~\ref{fig:output} illustrates several examples of the intermediate results obtained from our model, including the input faces, forward flow fields, neutral faces, backward flow fields, mean faces, and the faces reconstructed from their neutral ones. 
We demonstrate that the proposed method can handle face images with large variation in poses and can preserve facial attributes such as wearing glasses or beard.

Visualization of the facial motion flows presents both the head motion and the movement of facial muscles.
The neutral faces are deprived of facial motions in comparison to their original facial images.
Moreover, the mean faces obtained from different input images are almost identical to each other, which is in line with our assumption of identity invariance.

\subsection{Evaluation for Expression Representation}
\label{ExpRe}
Given the trained model, we investigate the learnt expression representation by evaluating the performance of its applications on expression recognition and head pose regression.
The goal is to verify whether the expression representation successfully encodes the information related to the facial motions and poses as our definition (\textit{i.e.} the expression factor contains all the variations between a face image and its corresponding neutral face of the same identity, including facial motions and head poses).
\Jia{We conduct \textit{linear-protocol} evaluation scheme for demonstrating the effectiveness of our method.}

\subsubsection{Expression Recognition}
Two datasets are used in the experiments of expression recognition, i.e. FER-2013~\cite{goodfellow2013challenges} and RAF-DB~\cite{li2017reliable}. FER-2013 dataset~\cite{goodfellow2013challenges} consists of 28,709 training and 3,589 testing images, while RAF-DB dataset consists of around 30K diverse facial images downloaded from the Internet. 
Please note that for the RAF-DB dataset, we follow the experimental setup as~\cite{li2017reliable} to particularly use the basic emotion subset of RAF-DB, which includes 12,271 training and 3,068 testing images.
For the evaluation scheme of linear-protocol, in order to directly verify the capacity of the expression features extracted by different models, we construct the \textit{linear} classifier upon the \textit{frozen} expression representations to perform the expression recognition, as in \cite{he2020momentum}. 
We follow the same procedure as \cite{he2020momentum} to train the linear layer (as the classifier) for 300 epochs, where the learning rate starts from 30 and decreases by a factor of 10 for every 80 epochs.
\Jia{The classifiers are trained by the SGD optimizer with cross-entropy objective and 256 batch size.}

%


%
The quantitative results shown in Table~\ref{tab:exp} demonstrate that the expression representation extracted from our proposed method is able to provide superior performance with respect to all the baselines.
%
\Jia{These results suggest that our proposed method can be used as a pretext task for expression recognition, where the rich information of facial expression is well learnt in a self-supervised manner.}

\begin{table}[t]
    \centering
    \begin{tabular}{l|c|c}
    \hline
        & FER-2013 & RAF-DB \\ \hline
        Method & Accuracy ($\%$) & Accuracy ($\%$)  \\ \hline
        \multicolumn{3}{l}{\textit{Fully supervised}} \\ \hline
        FSN \cite{zhao2018feature} & 67.60 & 81.10 \\ 
        ALT \cite{florea2019annealed} & 69.85 & 84.50 \\ \hline
        \multicolumn{3}{l}{\textit{Linear classification protocol}} \\ \hline
        LBP & 37.89 & 52.17 \\
        HoG & 45.47 & 63.53 \\
        FAb-Net \cite{wiles2018self} & 46.98 & 66.72 \\
        TCAE \cite{li2019self} & 45.05 & 65.32 \\
        BMVC'20 \cite{lu2020self} & 47.61 & 58.86 \\
        MoCo & 47.24 & 68.32 \\
        Ours & \textbf{48.76} & \textbf{71.01} \\ \hline
    \end{tabular}
    \caption{Evaluation on the task of expression classification based on the FER-2013 dataset~\cite{goodfellow2013challenges} and RAF-DB dataset~\cite{li2017reliable}.
    }
    \label{tab:exp}
\end{table}

\subsubsection{Regression of Head Pose}
Our definition indicates that the information of head poses would be also encoded into the expression representations. Obviously the calculated flow fields using the proposed method contain not only the local facial motion but also the global head motion, suggesting that our expression representation can also be used in the task of head pose regression. 
We adopt the 300W-LP~\cite{sagonas2013300} dataset and the AFLW2000~\cite{zhu2016face} dataset as the training and testing sets respectively, for experimenting the head pose regression. For the evaluation scheme of \textit{linear-protocol}, we construct a \textit{linear} regressor on top of the \textit{frozen} expression representations $E_{\textrm{exp}}$.
\Jia{The training runs for 300 epochs for the inear-protocol with SGD optimizer and batch size set to 16.}

As shown in Table~\ref{tab:pose}, for the linear-protocol evaluation scheme, the regressor based on our expression representations achieves 12.47 in terms of mean absolute error (MAE), which outperforms all the baselines.
These results demonstrate the effectiveness of our proposed method for well capturing the head pose information into expression representations.

\begin{table}[t]
    \centering
    \begin{tabular}{l|c|c|c|c}
    \hline
        Method & Yaw & Pitch & Roll & MAE  \\ \hline
        \multicolumn{5}{l}{\textit{Fully supervised}} \\ \hline
        FAN \cite{bulat2017far} & 6.36 & 12.3 & 8.71 & 9.12 \\
        FSA-Net \cite{yang2019fsa} & 5.27 & 6.71 & 5.28 & 5.75 \\ \hline
        \multicolumn{5}{l}{\textit{Linear regression protocol}} \\ \hline
        Dlib (68 points)~\cite{kazemi2014one} & 23.10 & 13.60 & \textbf{10.50} & 15.80 \\
        LBP & 23.58 & 14.86 & 16.36 & 18.27 \\
        HoG &  13.94 & 13.17 & 14.92 & 14.00 \\
        FAb-Net \cite{wiles2018self} & 13.92 & 13.25 & 14.51 & 13.89 \\
        TCAE \cite{li2019self} & 21.75 & 14.57 & 14.83 & 17.39 \\
        BMVC'20 \cite{lu2020self} & 22.06 & 13.50 & 15.14 & 16.90 \\        
        MoCo & 28.49 & 16.29 & 15.55 & 20.11\\ 
        Ours & \textbf{11.70} & \textbf{12.76} & 12.94 & \textbf{12.47} \\ \hline          
        
    \end{tabular}
    \caption{Evaluation on the task of head pose regression, where MAE stands for the mean absolute error.
    }
    \label{tab:pose}
\end{table}

\subsection{Evaluating Identity Representations}
\label{IdRe}
We also investigate the applications of identity representations learned by using the proposed method on the VoxCeleb dataset. 
Good performance of person recognition  demonstrates that our identity representations do contain rich information related to identities.

\subsubsection{Person Recognition}\label{sec:person_recog}
In this work we adopt LFW~\cite{huang2008labeled} and CPLFW \cite{CPLFWTech} dataset for the evaluation of person recognition, particularly on person verification.
The LFW dataset comprises of 13,233 face images from 5,749 identities and has 6,000 face pairs for evaluating person verification. 
The CPLFW dataset is similar to LFW but includes larger head pose variation.
We \textit{directly} extract the identity representations for all of the images in the face pairs from two datasets by using the encoder $E_{\textrm{id}}$ and then compute the cosine similarity between identity representations of each pair of face images.
Please note that, the features from baselines (\textit{i.e.} LBP, HoG, and MoCo) are also \textit{directly} applied to perform verification for a fair comparison.
As shown in Table~\ref{tab:id}, our identity representations can achieve 73.72\% in accuracy on LFW, which outperforms the unsupervised state-of-the-art method~\cite{datta2018unsupervised}.

\begin{table}[t]
    \centering
    \begin{tabular}{l|c|c}
    \hline
         &  LFW & CPLFW  \\ \hline
        Method & Accuracy(\%) & Accuracy(\%) \\ \hline
        \multicolumn{3}{l}{\textit{Fully supervised}} \\ \hline  
        VGG-Face~\cite{parkhi2015deep} & 98.95 &  84.00 \\
        SphereFace~\cite{liu2017sphereface} & 99.42 & 81.40\\
        ArcFace~\cite{deng2019arcface} & 99.53 & 92.08 \\ \hline   
        \multicolumn{3}{l}{\textit{Unsupervised or hand-crafted features}} \\ \hline
        VGG~\cite{datta2018unsupervised} & 71.48 & - \\         
        LBP & 56.90 & 51.50 \\
        HoG & 62.73 & 51.73 \\
        MoCo & 65.88 & 55.12\\
        Ours & \textbf{73.72} & \textbf{58.52} \\ \hline
    \end{tabular}
    \caption{Evaluation on the task of person recognition based on the LFW~\cite{huang2008labeled} and CPLFW~\cite{CPLFWTech} dataset. 
    We compare the performance of state-of-the-art methods in both supervised and unsupervised categories.
    }
    \label{tab:id}
\end{table}

\subsection{Frontalization}
\begin{figure}[t]
    \centering
    \includegraphics[clip,trim= 0cm 13.4cm 15.1cm 0 cm,width=0.9\columnwidth]{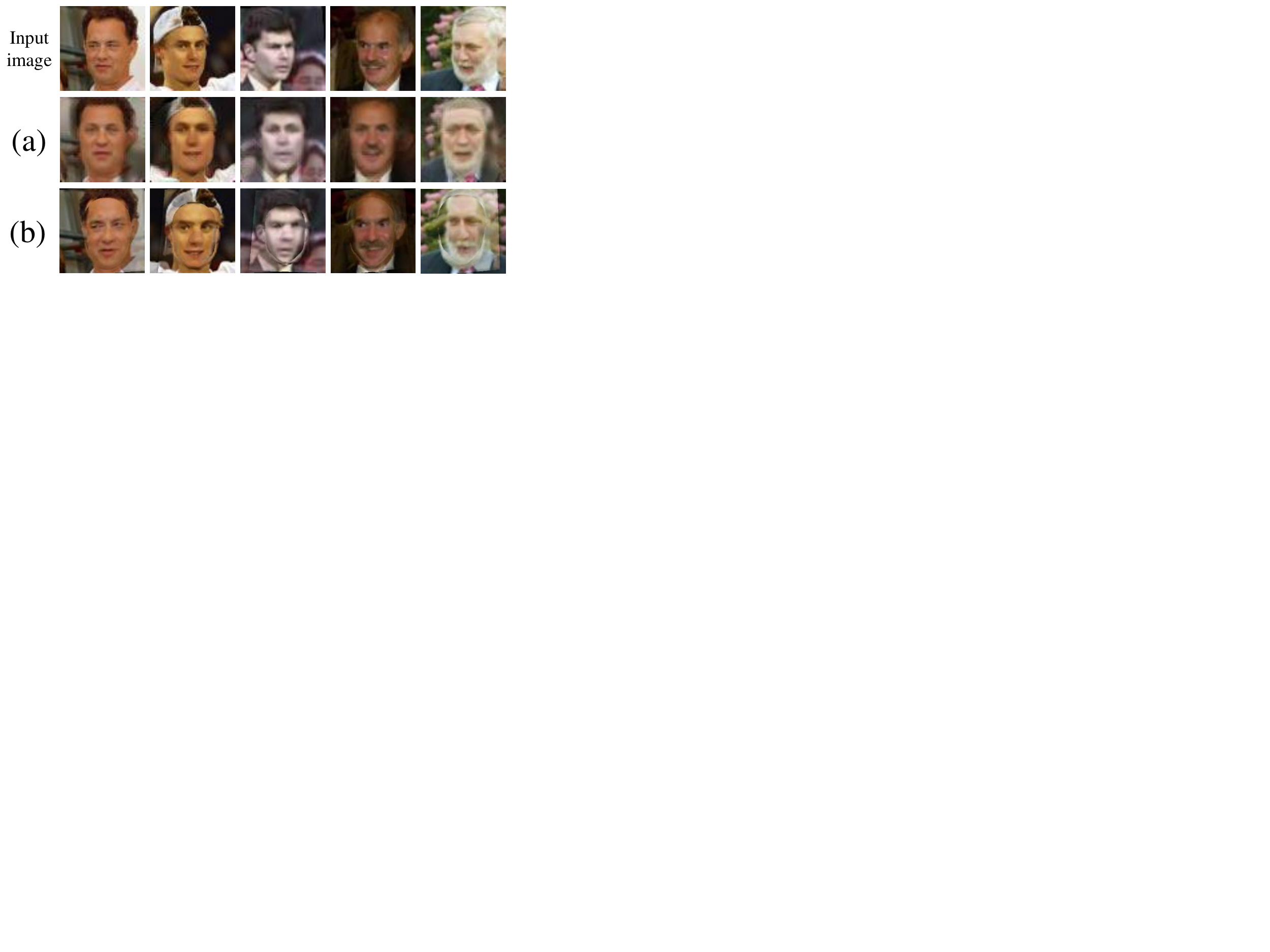}
    \caption{The frontalization results from (a) the proposed method and (b) the method in \cite{hassner2015effective}. These results clearly demonstrates the capacity of frontalization for facial images with various poses by using our method.}
    \label{fig:frontalization}
\end{figure}

Frontalization is the process of synthesizing the frontal facing view of a single facial image.
In this work, there are two ways to obtain the neutral face in the frontal view: de-expression and re-identity.
The de-expression operation removes the head motion and facial expressions from facial images and thus generates the neutral faces with frontal view.
On the other hand, the re-identity operation recovers the neutral face by adding the identity to the mean face which is already in frontal view.
As shown in Figure~\ref{fig:frontalization}(a), the proposed method is able to synthesize the neutral faces from the facial images with various poses by the de-expression operation.
The input images are from the LFW dataset~\cite{huang2008labeled} which are never seen during the training of our model. 
We also show a state-of-the-art approach in Figure~\ref{fig:frontalization}(b) which additionally uses facial landmarks~\cite{hassner2015effective} for qualitative comparison.

We notice that the synthesized images from the proposed method are a little bit blurry, we hypothesize that it might be caused by the plenty blurry training images in the Voxceleb dataset. We believe that further improvements can be obtained by using other high-quality datasets.

\subsection{Image-to-image Translation}
\label{ItoI}
\begin{figure}[t]
    \centering
    \includegraphics[clip,trim= 0cm 7.5cm 14.1cm 0 cm,width=0.9\columnwidth]{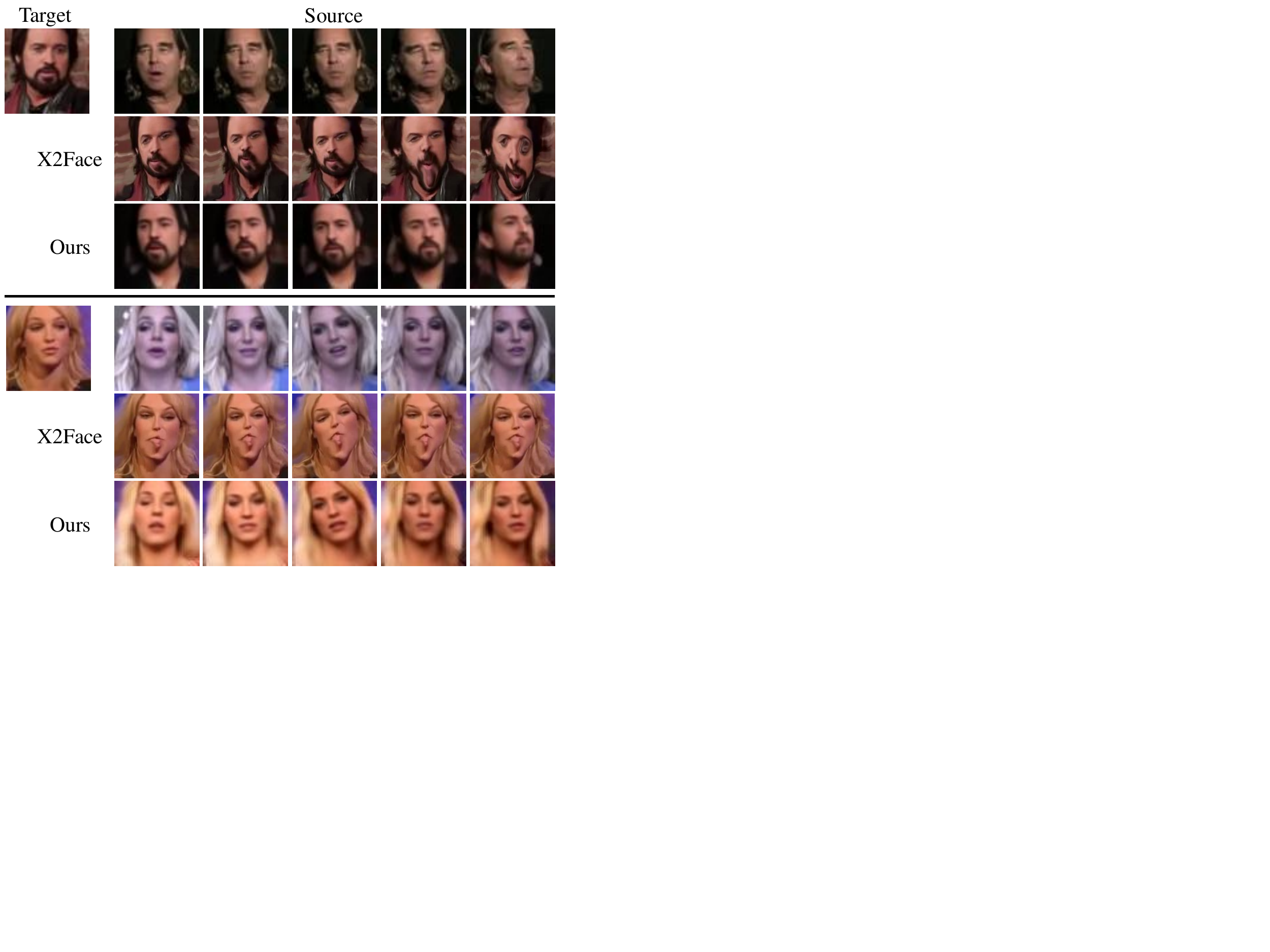}
    \caption{Example results of the proposed method on image-to-image translation in comparison to the  X2Face~\cite{wiles2018x2face} baseline. X2Face shows artifacts when performing large pose transfer. Notice that the proposed method does not include adversarial training to disentangle facial motion and to improve image quality.
    }
    \label{fig:trans}
\end{figure}

The proposed model can naturally be used to perform image-to-image translation by transferring the facial motion of the source image into the target one. 
To this end, we simply calculate and then apply the backward flow field of the source image to warp the neutral face of the target image via the re-expression operation.
As shown in Figure~\ref{fig:trans}, our method can transfer the head pose and expression from the source to the target without noticeable artifacts.
On the other hand, the results of X2Face method~\cite{wiles2018x2face} reveal visible artifacts when the pose difference between source and target is large.

\section{Conclusions}
In this work, we propose novel cycle-consistency constraints for disentangling of identity and expression representations from a single facial image, that is, facial motion cycle-consistency and the identity cycle-consistency.
The proposed model can be trained in an unsupervised manner by superimposing the proposed cycle-consistency constraints.
We perform extensive qualitative and quantitative evaluations on multiple datasets to demonstrate the efficacy of our proposed method on learning disentangled facial representations.
These representations contain rich and distinct information of identity and expression, and can be used to facilitate a variety of applications, such as facial expression recognition, head pose estimation, person recognition, frontalization, and the image-to-image translation.

\vspace{1em}
\noindent \textbf{Acknowledgement.}
This project is supported by MOST 108-2221-E-009-066-MY3, MOST 110-2636-E-009-001, and MOST 110-2634-F-009-018. We are grateful to
the National Center for High-performance Computing for providing
computing services and facilities.

{\small
\bibliographystyle{ieee_fullname}
\bibliography{egpaper_for_review.bbl}

\begin{thebibliography}{10}\itemsep=-1pt

\bibitem{black1996robust}
Michael~J Black and Paul Anandan.
\newblock The robust estimation of multiple motions: Parametric and
  piecewise-smooth flow fields.
\newblock {\em Computer vision and image understanding}, 63(1):75--104, 1996.

\bibitem{blanz1999morphable}
Volker Blanz, Thomas Vetter, et~al.
\newblock A morphable model for the synthesis of 3d faces.
\newblock In {\em ACM Transactions on Graphics (TOG)}, 1999.

\bibitem{bulat2017far}
Adrian Bulat and Georgios Tzimiropoulos.
\newblock How far are we from solving the 2d \& 3d face alignment problem?(and
  a dataset of 230,000 3d facial landmarks).
\newblock In {\em IEEE International Conference on Computer Vision (ICCV)},
  2017.

\bibitem{chu20143d}
Baptiste Chu, Sami Romdhani, and Liming Chen.
\newblock 3d-aided face recognition robust to expression and pose variations.
\newblock In {\em IEEE Conference on Computer Vision and Pattern Recognition
  (CVPR)}, 2014.

\bibitem{Chung18b}
J.~S. Chung, A. Nagrani, and A. Zisserman.
\newblock Voxceleb2: Deep speaker recognition.
\newblock In {\em INTERSPEECH}, 2018.

\bibitem{dalal2005histograms}
Navneet Dalal and Bill Triggs.
\newblock Histograms of oriented gradients for human detection.
\newblock In {\em IEEE Conference on Computer Vision and Pattern Recognition
  (CVPR)}, 2005.

\bibitem{datta2018unsupervised}
Samyak Datta, Gaurav Sharma, and CV Jawahar.
\newblock Unsupervised learning of face representations.
\newblock In {\em IEEE International Conference on Automatic Face \& Gesture
  Recognition (FG)}, 2018.

\bibitem{deng2019arcface}
Jiankang Deng, Jia Guo, Niannan Xue, and Stefanos Zafeiriou.
\newblock Arcface: Additive angular margin loss for deep face recognition.
\newblock In {\em IEEE Conference on Computer Vision and Pattern Recognition
  (CVPR)}, 2019.

\bibitem{ekman1978facial}
Paul Ekman and Wallace~V Friesen.
\newblock {\em Facial action coding system: Investigator's guide}.
\newblock Consulting Psychologists Press, 1978.

\bibitem{florea2019annealed}
Corneliu Florea, Laura Florea, Mihai-Sorin Badea, Constantin Vertan, and Andrei
  Racoviteanu.
\newblock Annealed label transfer for face expression recognition.
\newblock In {\em British Machine Vision Conference (BMVC)}, 2019.

\bibitem{gatys2016image}
Leon~A Gatys, Alexander~S Ecker, and Matthias Bethge.
\newblock Image style transfer using convolutional neural networks.
\newblock In {\em IEEE Conference on Computer Vision and Pattern Recognition
  (CVPR)}, 2016.

\bibitem{goodfellow2013challenges}
Ian~J Goodfellow, Dumitru Erhan, Pierre~Luc Carrier, Aaron Courville, Mehdi
  Mirza, Ben Hamner, Will Cukierski, Yichuan Tang, David Thaler, Dong-Hyun Lee,
  et~al.
\newblock Challenges in representation learning: A report on three machine
  learning contests.
\newblock In {\em International conference on neural information processing},
  2013.

\bibitem{hasselmo1989role}
Michael~E Hasselmo, Edmund~T Rolls, and Gordon~C Baylis.
\newblock The role of expression and identity in the face-selective responses
  of neurons in the temporal visual cortex of the monkey.
\newblock {\em Behavioural Brain Research}, 1989.

\bibitem{hassner2015effective}
Tal Hassner, Shai Harel, Eran Paz, and Roee Enbar.
\newblock Effective face frontalization in unconstrained images.
\newblock In {\em IEEE Conference on Computer Vision and Pattern Recognition
  (CVPR)}, 2015.

\bibitem{haxby2000distributed}
James~V Haxby, Elizabeth~A Hoffman, and M~Ida Gobbini.
\newblock The distributed human neural system for face perception.
\newblock {\em Trends in Cognitive Sciences}, 2000.

\bibitem{he2020momentum}
Kaiming He, Haoqi Fan, Yuxin Wu, Saining Xie, and Ross Girshick.
\newblock Momentum contrast for unsupervised visual representation learning.
\newblock In {\em IEEE Conference on Computer Vision and Pattern Recognition
  (CVPR)}, 2020.

\bibitem{huang2008labeled}
Gary~B. Huang, Manu Ramesh, Tamara Berg, and Erik Learned-Miller.
\newblock Labeled faces in the wild: A database for studying face recognition
  in unconstrained environments.
\newblock Technical report, University of Massachusetts, Amherst, 2007.

\bibitem{huang2017arbitrary}
Xun Huang and Serge Belongie.
\newblock Arbitrary style transfer in real-time with adaptive instance
  normalization.
\newblock In {\em IEEE International Conference on Computer Vision (ICCV)},
  2017.

\bibitem{hur2017mirrorflow}
Junhwa Hur and Stefan Roth.
\newblock Mirrorflow: Exploiting symmetries in joint optical flow and occlusion
  estimation.
\newblock In {\em IEEE International Conference on Computer Vision (ICCV)},
  2017.

\bibitem{Jiang_2019_CVPR}
Zi-Hang Jiang, Qianyi Wu, Keyu Chen, and Juyong Zhang.
\newblock Disentangled representation learning for 3d face shape.
\newblock In {\em IEEE Conference on Computer Vision and Pattern Recognition
  (CVPR)}, 2019.

\bibitem{johnson2016perceptual}
Justin Johnson, Alexandre Alahi, and Li Fei-Fei.
\newblock Perceptual losses for real-time style transfer and super-resolution.
\newblock In {\em European Conference on Computer Vision (ECCV)}, pages
  694--711. Springer, 2016.

\bibitem{kazemi2014one}
Vahid Kazemi and Josephine Sullivan.
\newblock One millisecond face alignment with an ensemble of regression trees.
\newblock In {\em IEEE Conference on Computer Vision and Pattern Recognition
  (CVPR)}, 2014.

\bibitem{li2017reliable}
Shan Li, Weihong Deng, and JunPing Du.
\newblock Reliable crowdsourcing and deep locality-preserving learning for
  expression recognition in the wild.
\newblock In {\em IEEE Conference on Computer Vision and Pattern Recognition
  (CVPR)}, 2017.

\bibitem{li2019self}
Yong Li, Jiabei Zeng, Shiguang Shan, and Xilin Chen.
\newblock Self-supervised representation learning from videos for facial action
  unit detection.
\newblock In {\em IEEE Conference on Computer Vision and Pattern Recognition
  (CVPR)}, 2019.

\bibitem{liu2017sphereface}
Weiyang Liu, Yandong Wen, Zhiding Yu, Ming Li, Bhiksha Raj, and Le Song.
\newblock Sphereface: Deep hypersphere embedding for face recognition.
\newblock In {\em IEEE Conference on Computer Vision and Pattern Recognition
  (CVPR)}, 2017.

\bibitem{lu2020self}
Liupei Lu, Leili Tavabi, and Mohammad Soleymani.
\newblock Self-supervised learning for facial action unit recognition through
  temporal consistency.
\newblock In {\em British Machine Vision Conference (BMVC)}, 2020.

\bibitem{lu2018attribute}
Yongyi Lu, Yu-Wing Tai, and Chi-Keung Tang.
\newblock Attribute-guided face generation using conditional cyclegan.
\newblock In {\em European Conference on Computer Vision (ECCV)}, 2018.

\bibitem{mase1991recognition}
Kenji Mase.
\newblock Recognition of facial expression from optical flow.
\newblock {\em IEICE TRANSACTIONS on Information and Systems}, 1991.

\bibitem{Nagrani17}
A. Nagrani, J.~S. Chung, and A. Zisserman.
\newblock Voxceleb: a large-scale speaker identification dataset.
\newblock In {\em Annual Conference of the International Speech Communication
  Association (INTERSPEECH)}, 2017.

\bibitem{ojala2002multiresolution}
Timo Ojala, Matti Pietikainen, and Topi Maenpaa.
\newblock Multiresolution gray-scale and rotation invariant texture
  classification with local binary patterns.
\newblock {\em IEEE Transactions on Pattern Analysis and Machine Intelligence
  (TPAMI)}, 24(7):971--987, 2002.

\bibitem{parkhi2015deep}
Omkar~M Parkhi, Andrea Vedaldi, and Andrew Zisserman.
\newblock Deep face recognition.
\newblock 2015.

\bibitem{pumarola2018ganimation}
Albert Pumarola, Antonio Agudo, Aleix~M Martinez, Alberto Sanfeliu, and
  Francesc Moreno-Noguer.
\newblock Ganimation: Anatomically-aware facial animation from a single image.
\newblock In {\em European Conference on Computer Vision (ECCV)}, 2018.

\bibitem{sagonas2013300}
Christos Sagonas, Georgios Tzimiropoulos, Stefanos Zafeiriou, and Maja Pantic.
\newblock 300 faces in-the-wild challenge: The first facial landmark
  localization challenge.
\newblock In {\em IEEE International Conference on Computer Vision Workshops},
  2013.

\bibitem{shen2017learning}
Wei Shen and Rujie Liu.
\newblock Learning residual images for face attribute manipulation.
\newblock In {\em IEEE Conference on Computer Vision and Pattern Recognition
  (CVPR)}, 2017.

\bibitem{shu2018deforming}
Zhixin Shu, Mihir Sahasrabudhe, Riza Alp~Guler, Dimitris Samaras, Nikos
  Paragios, and Iasonas Kokkinos.
\newblock Deforming autoencoders: Unsupervised disentangling of shape and
  appearance.
\newblock In {\em European Conference on Computer Vision (ECCV)}, 2018.

\bibitem{shu2017neural}
Zhixin Shu, Ersin Yumer, Sunil Hadap, Kalyan Sunkavalli, Eli Shechtman, and
  Dimitris Samaras.
\newblock Neural face editing with intrinsic image disentangling.
\newblock In {\em IEEE Conference on Computer Vision and Pattern Recognition
  (CVPR)}, 2017.

\bibitem{simonyan2014very}
Karen Simonyan and Andrew Zisserman.
\newblock Very deep convolutional networks for large-scale image recognition.
\newblock {\em ArXiv:1409.1556}, 2014.

\bibitem{sirovich1987low}
Lawrence Sirovich and Michael Kirby.
\newblock Low-dimensional procedure for the characterization of human faces.
\newblock {\em Journal of the Optical Society of America A}, 1987.

\bibitem{sun2018pwc}
Deqing Sun, Xiaodong Yang, Ming-Yu Liu, and Jan Kautz.
\newblock Pwc-net: Cnns for optical flow using pyramid, warping, and cost
  volume.
\newblock In {\em IEEE Conference on Computer Vision and Pattern Recognition
  (CVPR)}, 2018.

\bibitem{tran2017disentangled}
Luan Tran, Xi Yin, and Xiaoming Liu.
\newblock Disentangled representation learning gan for pose-invariant face
  recognition.
\newblock In {\em IEEE Conference on Computer Vision and Pattern Recognition
  (CVPR)}, 2017.

\bibitem{wiles2018self}
Olivia Wiles, A Koepke, and Andrew Zisserman.
\newblock Self-supervised learning of a facial attribute embedding from video.
\newblock In {\em British Machine Vision Conference (BMVC)}, 2018.

\bibitem{wiles2018x2face}
Olivia Wiles, A Sophia~Koepke, and Andrew Zisserman.
\newblock X2face: A network for controlling face generation using images,
  audio, and pose codes.
\newblock In {\em European Conference on Computer Vision (ECCV)}, 2018.

\bibitem{winston2004fmri}
Joel~S Winston, RNA Henson, Miriam~R Fine-Goulden, and Raymond~J Dolan.
\newblock fmri-adaptation reveals dissociable neural representations of
  identity and expression in face perception.
\newblock {\em Journal of Neurophysiology}, 2004.

\bibitem{xing2019unsupervised}
Xianglei Xing, Tian Han, Ruiqi Gao, Song-Chun Zhu, and Ying~Nian Wu.
\newblock Unsupervised disentangling of appearance and geometry by deformable
  generator network.
\newblock In {\em IEEE Conference on Computer Vision and Pattern Recognition
  (CVPR)}, 2019.

\bibitem{yang2018facial}
Huiyuan Yang, Umur Ciftci, and Lijun Yin.
\newblock Facial expression recognition by de-expression residue learning.
\newblock In {\em IEEE Conference on Computer Vision and Pattern Recognition
  (CVPR)}, 2018.

\bibitem{yang2019fsa}
Tsun-Yi Yang, Yi-Ting Chen, Yen-Yu Lin, and Yung-Yu Chuang.
\newblock Fsa-net: Learning fine-grained structure aggregation for head pose
  estimation from a single image.
\newblock In {\em IEEE Conference on Computer Vision and Pattern Recognition
  (CVPR)}, 2019.

\bibitem{zakharov2019few}
Egor Zakharov, Aliaksandra Shysheya, Egor Burkov, and Victor Lempitsky.
\newblock Few-shot adversarial learning of realistic neural talking head
  models.
\newblock In {\em IEEE International Conference on Computer Vision (ICCV)},
  2019.

\bibitem{zhang2018joint}
Feifei Zhang, Tianzhu Zhang, Qirong Mao, and Changsheng Xu.
\newblock Joint pose and expression modeling for facial expression recognition.
\newblock In {\em IEEE Conference on Computer Vision and Pattern Recognition
  (CVPR)}, 2018.

\bibitem{zhang2018unsupervised}
Yuting Zhang, Yijie Guo, Yixin Jin, Yijun Luo, Zhiyuan He, and Honglak Lee.
\newblock Unsupervised discovery of object landmarks as structural
  representations.
\newblock In {\em IEEE Conference on Computer Vision and Pattern Recognition
  (CVPR)}, 2018.

\bibitem{zhao2018feature}
Shuwen Zhao, Haibin Cai, Honghai Liu, Jianhua Zhang, and Shengyong Chen.
\newblock Feature selection mechanism in cnns for facial expression
  recognition.
\newblock In {\em British Machine Vision Conference (BMVC)}, 2018.

\bibitem{CPLFWTech}
T. Zheng and W. Deng.
\newblock Cross-pose lfw: A database for studying cross-pose face recognition
  in unconstrained environments.
\newblock Technical Report 18-01, Beijing University of Posts and
  Telecommunications, February 2018.

\bibitem{zhu2016face}
Xiangyu Zhu, Zhen Lei, Xiaoming Liu, Hailin Shi, and Stan~Z Li.
\newblock Face alignment across large poses: A 3d solution.
\newblock In {\em IEEE Conference on Computer Vision and Pattern Recognition
  (CVPR)}, 2016.

\end{thebibliography}
}

\end{document}